\documentclass[runningheads]{llncs}
\usepackage{graphicx}
\usepackage{algorithm}
\usepackage{algorithmic}
\usepackage{enumerate}
\usepackage{array}
\usepackage{booktabs}
\usepackage{multirow}
\begin{document}
\title{Deep Dual Pyramid Network for Barcode Segmentation using Barcode-30k Database}
%
%
\author{Qijie Zhao$^{1\dagger}$, Feng Ni$^{1\dagger}$, Yang Song$^{2}$, Yongtao Wang$^1$\thanks{Corresponding Author} , Zhi Tang$^1$\\}

\institute{$^1$Institute of Computer Science and Technology, Peking University\\
$^2$Tsinghua University\\
${^\dagger}$indicates equal contribution.\\
\tt\small \{zhaoqijie,nifeng,wyt,tangzhi\}@pku.edu.cn, song-y16@mails.tsinghua.edu.cn\\
}
\maketitle              

\begin{abstract}
Digital signs(such as barcode or QR code) are widely used in our daily life, and for many applications, we need to localize them on images. However, difficult cases such as targets with small scales, half-occlusion, shape deformation and large illumination changes cause challenges for conventional methods. In this paper, we address this problem by producing a large-scale dataset and adopting a deep learning based semantic segmentation approach. Specifically, a synthesizing method was proposed to generate well-annotated images containing barcode and QR code labels, which contributes to largely decrease the annotation time. Through the synthesis strategy, we introduce a dataset that contains 30000 images with Barcode and QR code - Barcode-30k. Moreover, we further propose a dual pyramid structure based segmentation network - BarcodeNet, which is mainly formed with two novel modules, Prior Pyramid Pooling Module(P3M) and Pyramid Refine Module(PRM). We validate the effectiveness of BarcodeNet on the proposed synthetic dataset, and it yields the result of mIoU accuracy 95.36\% on validation set. Additional segmentation results of real images have shown that accurate segmentation performance is achieved.


\end{abstract}

\section{Introduction}

Digital sign is able to represent rich information of the attached product, such as its identification number and the production date. It usually has a relative regular shape in 2D images. Barcode and QR code, the most widely appeared digital signs in common life, are employed in factories and supermarkets massively. Using computer vision algorithms to help accurately detect, segment and recognize these digital signs will greatly promote the development of automation in these industries. With the powerful deep convolutional neural networks, the performance of most visual tasks has been greatly improved. For instance, deep learning methods \cite{Xie2016Aggregated,Shelhamer2017Fully,Zhao2017Pyramid,Chen2018DeepLab,Chen2015Semantic,Caesar2016Region,Dai2014Convolutional,ChenPSA17,abs-1802-02611,Yu2015Multi} have achieved a breakthrough in the field of semantic segmentation.

Images with barcode and QR code, we call them barcode images in this paper for simplicity. It's not easy to directly apply existing semantic segmentation methods (such as FCN \cite{Shelhamer2017Fully}, PSPNet \cite{Zhao2017Pyramid}), SegNet \cite{Badrinarayanan2017SegNet}, Deeplab \cite{Chen2018DeepLab} ,DeepLabv3 \cite{ChenPSA17} and so on)  to segment QR code and barcode pixels from barcode images due to the following two problems:

\begin{enumerate}
\item \textbf{No well-annotated dataset}: Barcode images of existing barcode dataset such as \cite{Zamberletti2010Neural,Wachenfeld2010Robust,Zamberletti2013Robust} are taken closer than 30cm and under clear lighting, while no annotated barcode images are taken at more common conditions, e.g., around 0.5 meter to 2 meters to the targets. Moreover, annotating pixel-level labels as supervised segmentation tasks need is very time-consuming, especially when the target barcode is even too small to be noticed by human eyes.
\item \textbf{No off-the-shelf deep models}: State-of-the-art networks, including PSPNet and DeepLabv3 \cite{ChenPSA17}, are mainly designed for irregular object segmentation, such as car and pedestrian, which mostly have soft borders. Digital signs, like QR codes and barcodes, always appear with regular shapes. Unlike objects with high-level semantics, digital signs contain plenty of structural texture information. However, no models take use of very shallow features for segmentation recently.
\end{enumerate}

\begin{figure}[t]
\centering
\includegraphics[scale=0.35]{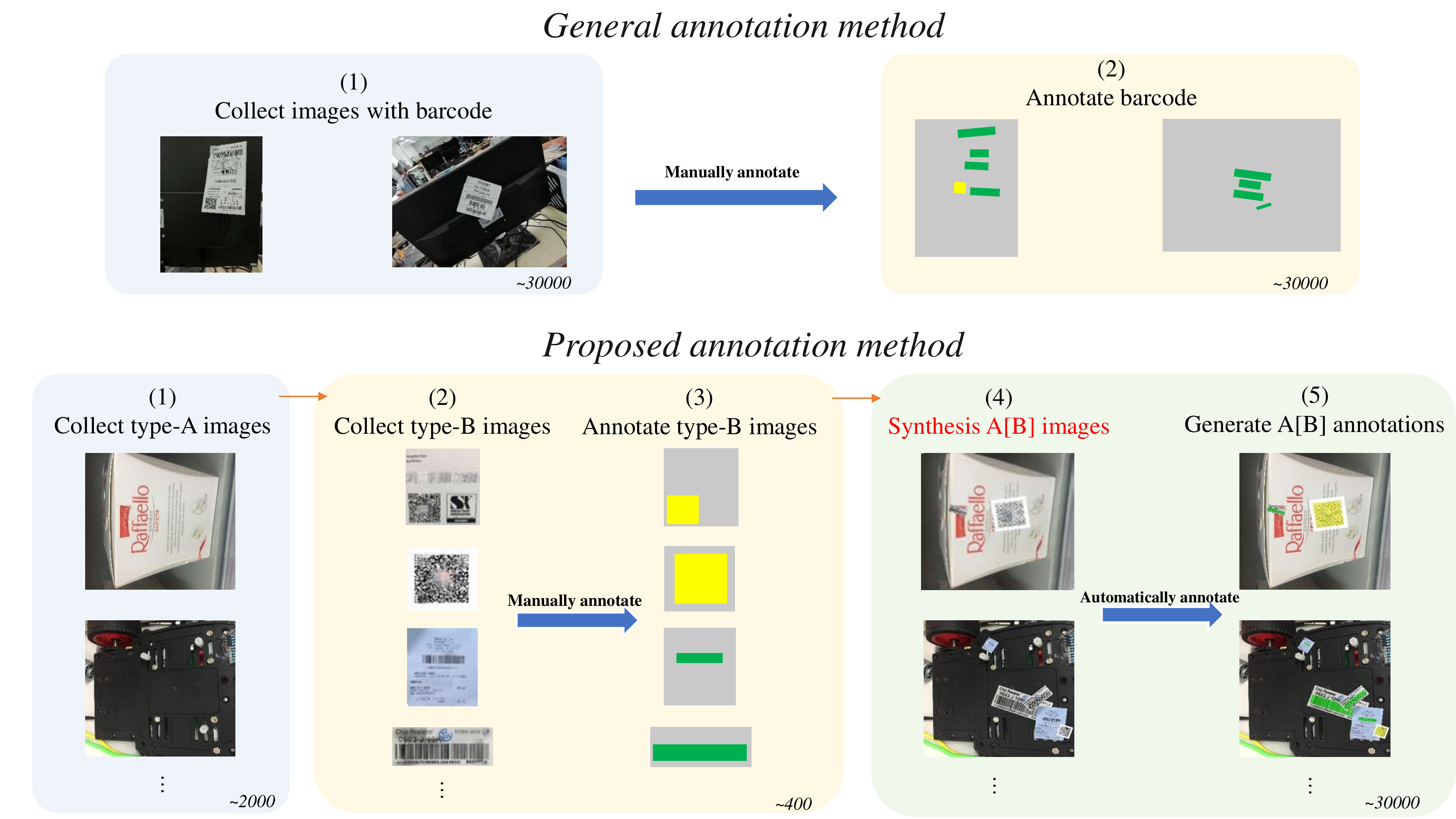}
\caption{Comparison of the annotation steps between general method and the method driven by our proposed synthesis strategy. The upper subfigure is general annotation method, the bottom subfigure is our proposed annotation steps. Synthesizing operation saves a lot of annotation time. \textbf{In this paper, the sensitive information on images is pixelated into the mosaic.}}
\label{fig:ss}
\end{figure}

To solve the above two problems, we first propose a method for synthesizing dataset as shown in Fig~\ref{fig:ss}. 
Specifically, we use around 2000 type-A images and around 400 type-B images to create a large dataset with $\sim$30000 barcode images(namely Barcode-30k). The synthetic barcode dataset Barcode-30k includes cases of one or more barcodes on images, as well as different scales, different rotation angles and different cropping ratios, which will help models improve the generalization abilities.

Considering the nature of digital signs, we designed a semantic segmentation network - BarcodeNet, based on PSPNet but more accurate than it for segmenting regular barcode and QR code. Our BarcodeNet introduces a dual pyramid structure to help learn both the rich semantic information and structural information from the feature maps. 
Specifically, the Prior Pyramid Pooling module(P3M) uses global prior information to guide local location distribution, which considers more details for large targets. And the other Pyramid Refine module(PRM) helps learn the shallow features of the images, aiming at refining more regularized boundaries for the segmented pixels. Also, shallow features can help recognize barcode images.

In the following content, we will introduce the related work of recent segmentation researches and existing barcode dataset in Section 2. And in Section 3, we describe the strategy to synthesize a barcode dataset. In Section 4, we introduce the proposed BarcodeNet. Then, experiments are shown in Section 5, discussing problems about barcodes in Section 6. The inference model and code of BarcodeNet will be publicly available soon.

\section{Related Work}
    \subsection{Traditional Segmentation methods}

Mostly based on the low-level visual cues of the pixel itself, traditional semantic segmentation approaches usually rely on domain knowledge to extract features and then apply them with post processing methods such as Texton Forests \cite{Johnson2008Semantic}, SVM \cite{Cortes1995Support} and Conditional Random Fields (CRFs) \cite{Lafferty2001Conditional}. However, for more challenging segmentation tasks, if no artificial information is provided, the segmentation performance will meet a significant limitation. As a result, they were quickly replaced by effective deep learning algorithms.

    \subsection{Deeplearning segmentation methods}

In the past few years, CNN-based networks\cite{Garcia2017A,Thoma2016A} have been proposed and make a breakthrough in the field of segmentation. Fully convolutional network (FCN) \cite{Shelhamer2017Fully}, which includes the implementation of deconvolution layer, upsampling and skip architecture to fuse the high-level and low-level features, is the pioneering work in the field of semantic segmentation. Dilated convolution \cite{Yu2015Multi} and DeepLab(v1,v2,v3,v3+) \cite{Chen2018DeepLab,Chen2015Semantic,ChenPSA17,abs-1802-02611} use subsequent convolution layers to expand the receptive field and maintain resolution. Other segmentation networks such as encoder-decoders \cite{Badrinarayanan2017SegNet}, region-based representations \cite{Caesar2016Region,Dai2014Convolutional}, and cascaded networks \cite{Lin2016RefineNet} also contribute to the trend of research. In order to achieve higher accuracy, several supporting techniques are merged, including ensembling features \cite{Chen2018DeepLab}, multi-stage training \cite{Shelhamer2017Fully}, additional training data from other datasets \cite{Chen2018DeepLab}, object proposals \cite{Noh2015Learning}, CRF-based post processing \cite{Xie2016Aggregated} and pyramid-based feature re-sampling\cite{Zhao2017Pyramid}. 


  \subsection{Barcode segmentation}

Research on segmentation algorithm of barcode in the past were mainly focusing on images taken within 30 centimeters and in each image there was only one no-overlapped complete barcode. \cite{Li2013Adaptive} proposed an adaptive segmentation method based on Mathematic Morphological, \cite{Liu2012Research} presents a segmentation algorithm for 2D color barcode image based on gradient features. Xu et al.\cite{Xu20112D}developed an approach for detecting blur 2D barcodes based on coded exposure algorithms. Li et al. \cite{Li2017Morphological} propose two 2-D barcode image segmentation algorithms under complex background.  Li and Zhao et al. \cite{Li2017Using} proposeda a cascaded strategy for accurate detection of 1D barcode with deep convolutional neural network.


%


\section{Synthesize Barcode Images}

In this section, we mainly introduce the synthesizing strategy for barcode images. Specifically, we describe how to collect and annotate images, and the detailed synthesizing steps in section 3.1. Then in Section 3.2, we introduce the Barcode-30k dataset, which is gained by above synthesis strategy. Note that, we overallly annotate these digital signs into 2 categories: \textit{Barcode} and \textit{QR code} , indicated as green and yellow mask as shown in Fig1.
  
  \subsection{Synthesis Strategy}
Digital signs need rare context information to localize because they are always pasted randomly on various backgrounds, which is available for us to generate similar images. To build a large-scale barcode dataset, we synthesize barcode images by affixing smaller images with barcode(name type-B images in Fig.1) to relative larger background images(name type-A images in Fig.1) following a series of transformation rules. What we should annotate manually is the type-B images, their pixels are easy to label due to the regular target areas.

In order to simulate, we set some basic rules to synthesize: Each synthesized image(denote as A[B]) contains at least one sampled type-B image; A[B] should contain less than 10 type-B imagest, half contain same type-B images and half contain different type-B images, the number of type-B images in one type-A image obedience to a poisson-distribution. Most importantly, type-B images should randomly be transformed by multiple operations, including half-cut, distorting, lighting-darken and made into low-resolution.
%

A more detailed description of the strategy is listed as follows: 

(1) For each background image of type-A, we randomly(50\%) select one type-B image or randomly(50\%) select $T$ type-B images as Selected type-B images(STB1). Then get a number $K$ that from Poisson-distribution($\lambda=2)$ and sample $K$ times type-B images from $STB1$ randomly with repeated samples, denote the $K$ type-B images as $STB2$.

(2) For each barcode image in $STB2$, apply a series of transformation steps: Resizing by the ratio from 0.2 to 5; Rotating by the ratio from $-45^{\circ}$ to $45^{\circ}$; Cropping a corner randomly(10\%) with a ratio from 0.5 to 1; lighting-darkening with a ratio from 1 to 0.5; Resizing twice(zoom in and zoom out) to make into low-resolution. After these, $STB2$ images are prepared.

(3) These $STB2$ images are randomly pasted on type-A image. Only one post processing step, any pixels that beyond the size range of the type-A image, directly crop off. By the way, the pixel-level labels of A[B] images are automatically generated within the pasting processing.



 \subsection{Barcode-30K dataset}
We generate Barcode-30k following 2 steps: collecting original type-A and type-B images, synthesizing the simulated data using the above synthetic strategy.

For type-A images, we have selected some representative places from indoor and outdoor situations, containing a diverse number of different scenes. As shown in Fig.3, in our daily life, supermarkets, laboratories, and offices that contain most of the indoor scenes, in which there are plenty of goods, devices, express delivery boxes, etc. Meanwhile streets, parking lots and parks cover most outdoor scenes, in which the most common things are billboards, cars, etc. The distance between camera and targets are mostly from half meters to 2 meters. As for type-B images, we mainly take images that have barcodes and QR codes with a close distance, so that bring convenience for annotation step.

Finally, we selected 2000 type-A images, and 400 type-B images which we have accurate annotated. Based on these images, with the above synthesis strategy, we synthesize a dataset that contains 30,000 Barcode images, called Barcode-30k. Some sampled images from Barcode-30k are shown in Fig.2, the barcodes are just similar to those posted by people who posted small advertisements. Barcode-30k not only contains large number of well-annotated images, but also is with enough diversity. It includes two kinds of label information: QR code and bar code. The synthesis process of Barcode-30k provides guidance for many segmentation tasks and detection tasks for regular targets.

\begin{figure}[t]
    \begin{minipage}[t]{0.5\linewidth}
    \centering
    \includegraphics[height=3cm,width=6cm]{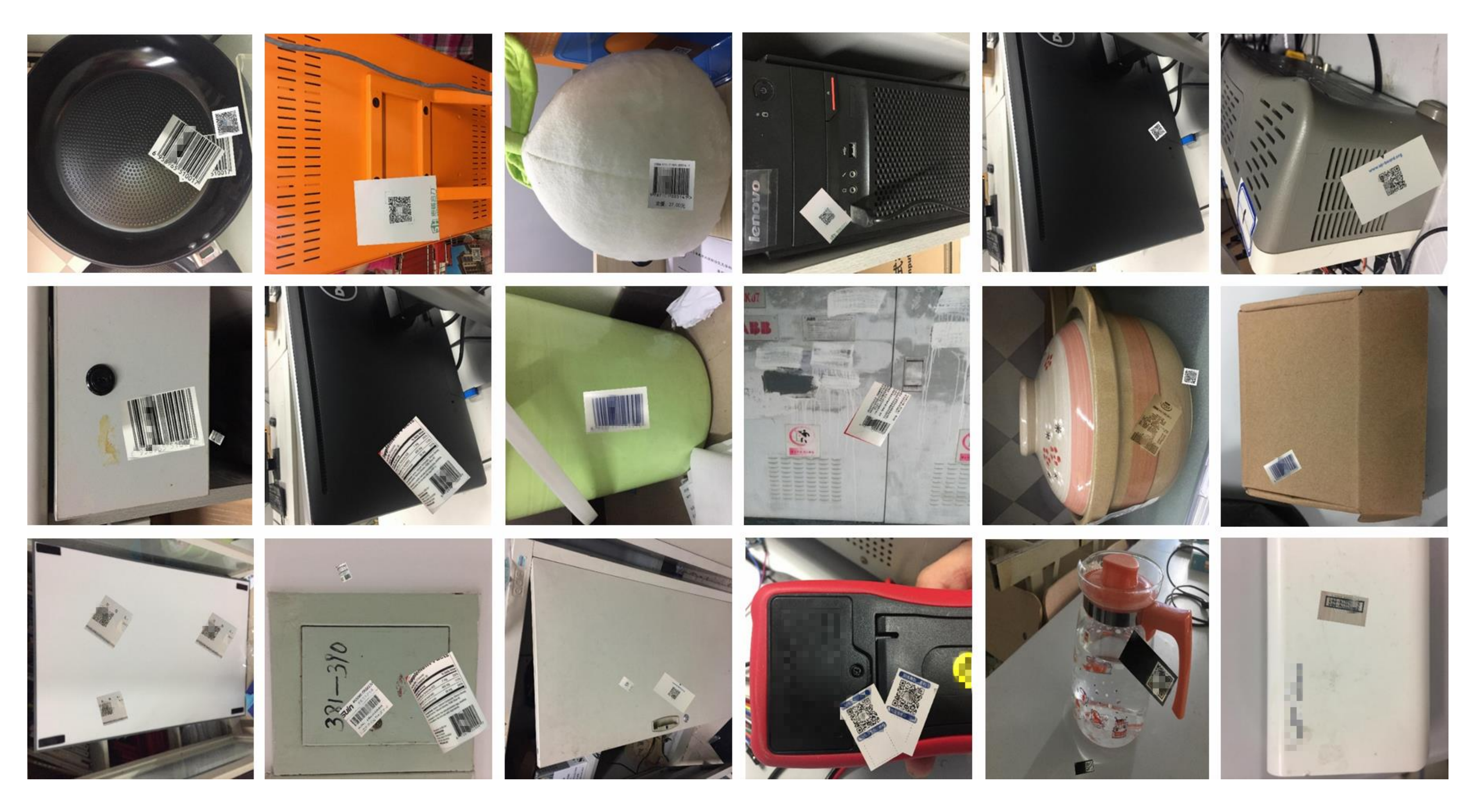}
    \caption{Examples of Barcode-30k.}   
     \label{fig:example}
    \end{minipage}
    \begin{minipage}[t]{0.5\linewidth}   
     \centering
    \includegraphics[height=3.0cm,width=5.2cm]{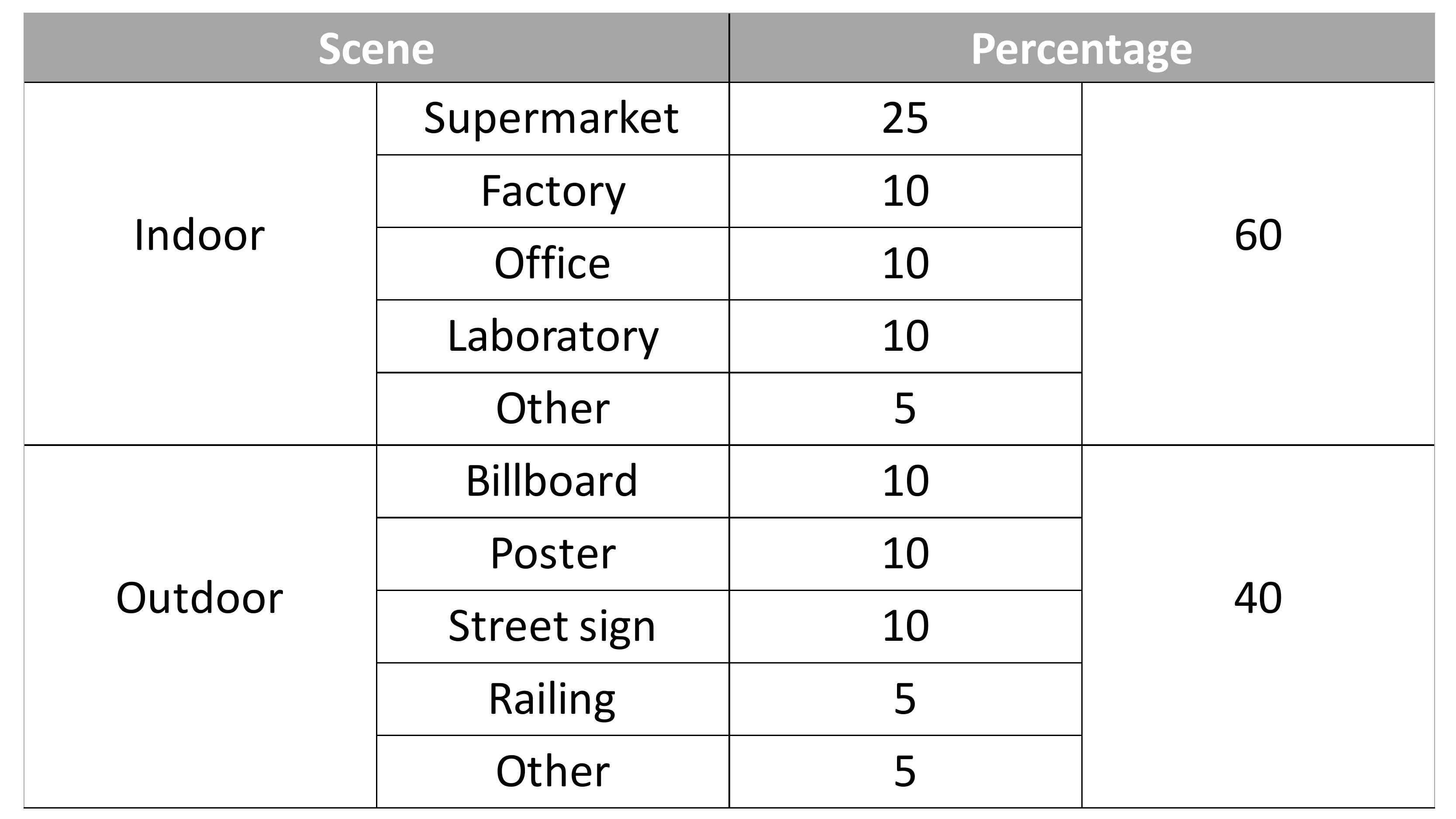}
    \caption{Distribution of type-A images.}
    \label{fig:distribution}
    \end{minipage}

\end{figure}


\section{BarcodeNet}

\begin{figure}[t]
\centering
\includegraphics[scale=0.35]{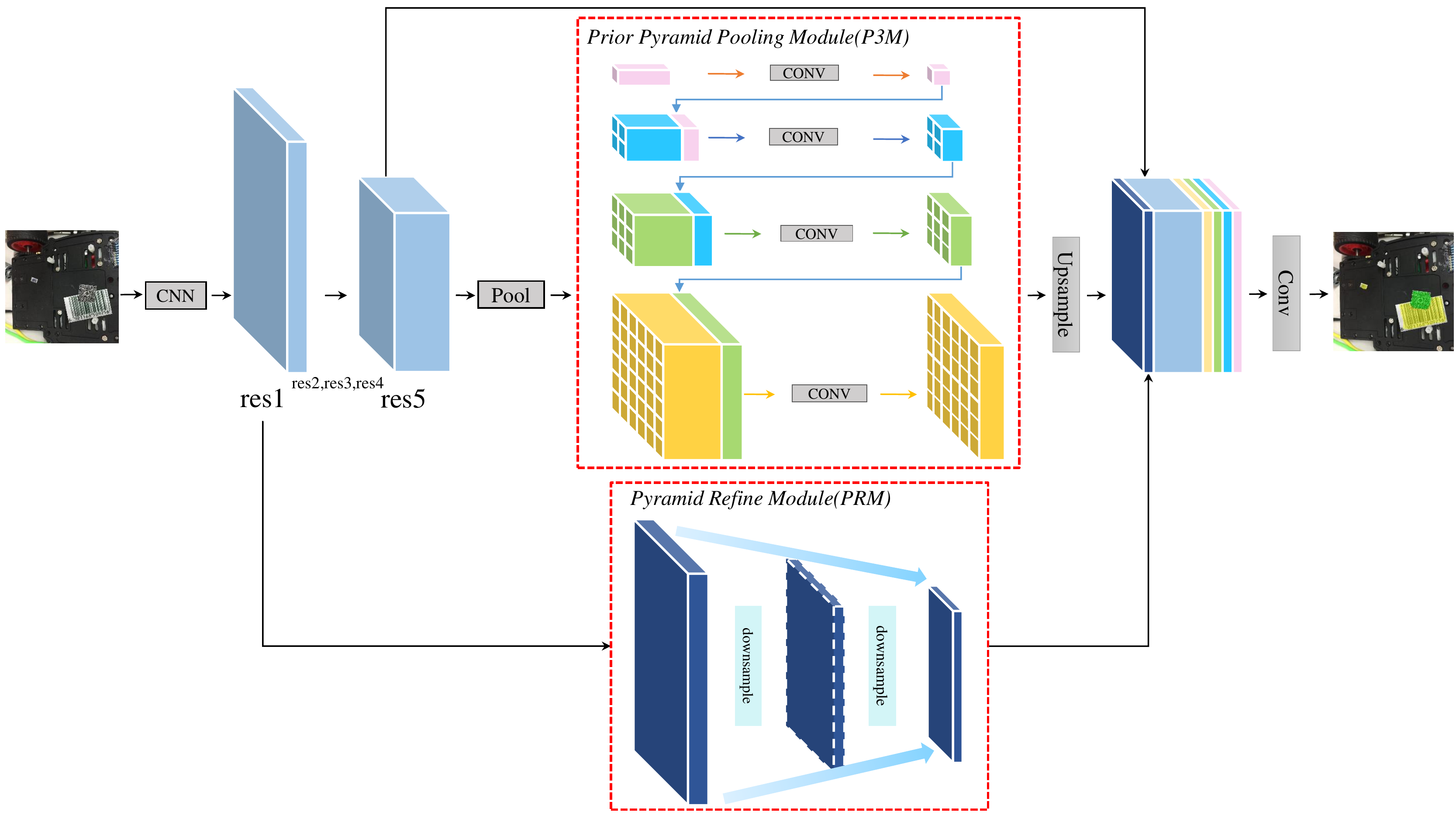}
\caption{Illustration of our framework, the Dual Pyramid Network - BarcodeNet.}
\label{fig:BarcodeNet}
\end{figure}

Refer to the overall network architecture shown in Fig~\ref{fig:BarcodeNet}. Similar to PSPNet\cite{Zhao2017Pyramid}, BarcodeNet obtains global and local pixel classification features by pooling multi-scale features based on Res5. In addition, the features of Res1 are extended to be utilized, because shallow features have stronger structural information, which is effective for apperceiving regular objects\cite{Zeiler2014Visualizing}. The backbone of BarcodeNet is ResNet101 \cite{He2016Deep} , more powerful to extract representative features compared with shallow networks, e.g. VGG16. BarcodeNet is mainly formed by two pyramid pooling modules, the top Prior Pyramid Pooling Module(P3M) and the bottom Pyramid Refine Module(PRM) shown in Fig.4. The P3M is designed for aggregating contextual information, the global average pooling operations based on multi-scale global features could learn global representative features for objects ranges among different scales. And the PRM is adept at using shallow features to learn regular geometrical boundaries. The two modules are described in detail in the following, and moreover, we also expound the effective optimization method for ResNet-FCN like network\cite{Zhao2017Pyramid}.

\textbf{Prior Pyramid Pooling Module(P3M):} Pyramid Pooling module(P2M) is introduced to learn global knowledge by pooling with multi-scale kernels\cite{Zhao2017Pyramid,He2015Spatial,Liu2015ParseNet}. However, no methods ever considered their relations, i.e., global guidance between each other. Considering the fact that feature maps in original pyramid pooling module(P2M) lack semantic information and all layers directly make predictions without guidance, resulting worse segmentations. Besides P2M, to further strengthen semantic information relationship between feature maps with different scales, we propose a hierarchical global prior, containing information with different scales and varying among different sub-regions. Specifically, we added another branch from the output feature map of upper pooling module to the input of next pooling module. As illustrated in Fig~\ref{fig:BarcodeNet}, the blue lines connecting adjacent two-layer features indicate the priority information we introduced. Then we directly upsample the feature maps of prior pyramid pooling module to generate the same size feature as the original feature map via bilinear interpolation.

%

\textbf{Pyramid Refine module(PRM):} Different from the general segmentation task, the barcode always have regular outlines, and therefore, the structural information of shallow feature maps is of great significance. This inspired us to utilize another pyramid module from the output of shallow features, as illustrated in Fig~\ref{fig:BarcodeNet}. To keep the structural features of shallow positions, we try to make PRM as simplistic as possible. In detail, two groups of Conv(3x3)+BN+ReLU are used continuously to construct PRM part, note that both Conv strides are set as 2 to match the size of Res5 feature. The Pyramid Refine Module(PRM) learns shallow features like edges, textures and streaks from feature Res1, which leads to refine out more regularized boundaries for the targets.

Finally, multiple kinds of features generated by P3M and PRM, and the raw Res5 feature are concatenated as the final representative feature. This powerful feature fusion system makes good use of global image-level priors, multi-scale features and structural information.

\section{Experiments}
In this section, we evaluate our approach BarcodeNet on the proposed dataset.
To measure the performance, we focused our experiments on mIoU(mean of Intersection over Union) percentage, which becomes a standard for semantic segmentation.
To compare different modules and different settings of our method, we conduct experiments on a mini subset of Barcode-30k - Barcode-3k, which is evenly sampled by the ratio of 1:10. Finally, to evaluate ultimate performance of BarcodeNet, we change to full Barcode-30k. Note that the training and validation ratio is split by 4:1 for both.

\begin{table}[b]
\caption{Segmentation results of different structures.}
\label{tab:segresults}
\centering
\begin{tabular}{p{2.5cm}<{\centering}|p{2.0cm}<{\centering}|p{1.7cm}<{\centering}|p{2cm}<{\centering}|p{1.1cm}<{\centering}|p{1.0cm}<{\centering}}\hline
\toprule
\multirow{1}{*}{Method} & \multirow{1}{*}{Backbone} & \multirow{1}{*}{Input size} & \multirow{1}{*}{Mean Acc} & \multirow{1}{*}{FPS}& \multirow{1}{*}{mIoU} \\ 
\hline
FCN-8s \cite{Shelhamer2017Fully} &VGG-16& 480 $\times$ 480 & 85.38 & 9.17 & 82.00 \\
PSPNet \cite{Zhao2017Pyramid} &ResNet-101& 473 $\times$ 473 & 92.15 & 5.88 & 89.23\\
PSPNet-P3M &ResNet-101& 473 $\times$ 473 & 93.26 & 5.63 & 90.36 \\
\textbf{BarcodeNet} &ResNet-101 & 473 $\times$ 473 & \textbf{94.24} & 5.23 & \textbf{91.63}\\
\bottomrule
\end{tabular}
\label{tab:segresults}
\end{table}

\subsection{Configurations of BarcodeNet}

We train models with end-to-end manners on 2 NVIDIA GTX1080Ti GPUs, optimized by synchronized SGD with a weight decay of 0.001 and momentum of 0.9. Giving a batch size of 8 and an initial learning rate of 0.001, we then reduce the learning rate down to 0.0001 when it comes to 50 epochs, and 0.00001 when 80 epochs, it can reach optimum performance at 90 epochs. We have adopted FCN-8s as our baseline, and selected PSPNet as the main comparison. To enhance performance, we pretrained all models using Cityscapes \cite{Cordts2016The}.

\subsection{Validate P3M and PRM}

The segmentation results of our methods are shown in Table~\ref{tab:segresults}. With our proposed BarcodeNet, the best mIoU results reaches 91.63\%. And compared to the baseline FCN-8s, BarcodeNet outperforming it by nearly 10\%. Specifically, by using Pyramid Pooling module that PSPNet introduced, the mIoU then improved by 7.23\% compared with FCN-8s baseline. As for P3M, we validate their contribution by evaluate PSPNet-P3M, which replace Pyramid Pooling module with Prior Pyramid Pooling Module. Notably, P3M improves by 1.1\%, with very little added modelweights. Since shallower layer feature contains more structural information in deep networks, PRM is specially designed for refining structural information. Using both P3M and PRM in PSPNet, it becomes our BarcodeNet. It improved the mIoU by 2.4\% compared with the original PSPNet, and 1.3\% compared with the PSPNet using P3M, which shows the importance of fusing refined structural information by PRM. more clearly visual comparisons can be seen in Fig~\ref{fig:results}. In addition, Barcode that trained on full Barcode-30k training set can achieve mIoU of 0.9536.

\begin{table}[htbp]
  \caption{PRM from different shallow features.}
  \label{tab:PRMloc}
  \centering
  \begin{tabular}{p{2cm}<{\centering}|p{2.8cm}<{\centering}|p{1.8cm}<{\centering}|p{2.0cm}<{\centering}|p{2.0cm}<{\centering}|p{1.2cm}<{\centering}}
    \toprule
    PRM feature & Layers &  Mean Acc & barcode IoU & QR code IoU & mIoU\\
    \hline
    Res5 & Encoder-Decoder & 93.09 & 83.25 & 89.32 & 90.65\\
    Res4 & Encoder-Decoder & 93.30 & 83.53 & 89.65 & 90.76\\
    Res3 & Encoder-Decoder & 93.63 & 83.76 & 89.93 & 90.94\\
    Res2 & Two Convs & 93.96 & 84.05 & 90.18 & 91.13\\
    \textbf{Res1} & \textbf{Two Convs} &\textbf{94.24}  &\textbf{84.72} & \textbf{90.73} &\textbf{91.63}\\
  \bottomrule
  \end{tabular}
  \label{tab:PRMloc}  
\end{table}

  \subsection{Evaluate the structure of PRM}
Fix the P3M, we conduct experiments to verify different PRM modules. The role of PRM module in BarcodeNet is to aggregate more shallow features to the final fusion features as refinement of the regular position at boundaries. The PRM module obviously has different structural settings, and it can also learn shallow information from features from different depths. Shown in Tab.2, we compare the guidance knowledge from different depths as shallow feature(Res1,2,3,4,5). For Res1 and Res2, we connect the two conv layers directly and set the stride parameter to 2 to get the same scale with final features. For Res3,4 and 5, we use Encode-Decode structure to connect the features into the Conv layer and the Deconv layer. Experiments show that the guidance of Res1 feature can bring more improvement, and from a unified perspective, the shallower layer gets better results.

\begin{table}[b]
  \caption{PRM of different operations.}
  \label{tab:PRMstr}
  \centering
  \begin{tabular}{p{2cm}<{\centering}|p{2.8cm}<{\centering}|p{1.8cm}<{\centering}|p{2.0cm}<{\centering}|p{2.0cm}<{\centering}|p{1.2cm}<{\centering}}
    \toprule
    PRM feature & Layers & Mean Acc  & barcode IoU & QR code IoU  & mIoU\\
    \hline
    \multirow{3}{*}{Res1} & Maxpooling(s=4) & \multirow{1}{*}{93.55} & \multirow{1}{*}{83.75} & \multirow{1}{*}{90.00} & \multirow{1}{*}{91.06}\\
    & Avgpooling(s=4) & \multirow{1}{*}{93.94} & \multirow{1}{*}{84.12} & \multirow{1}{*}{90.31} & \multirow{1}{*}{91.28}\\
    & \textbf{Two Convs(s=2)} & \multirow{1}{*}{\textbf{94.24}} & \multirow{1}{*}{\textbf{84.72}} & \multirow{1}{*}{\textbf{90.73}} & \multirow{1}{*}{\textbf{91.63}}\\
  \bottomrule
  \end{tabular}
  \label{tab:PRMstr}
\end{table} 
In addition, we also validate how to use shallow information, whether by directly pooling or through the convolutional layers to learn the features we need. Tab.3 compares these different operations. The experimental results show that the two-layer convolution PRM structure works better, allowing the final BarcodeNet to reach mIoU of 91.63 .

  \subsection{More visual comparisons}

\begin{figure}[t]
\centering
\includegraphics[scale=0.36]{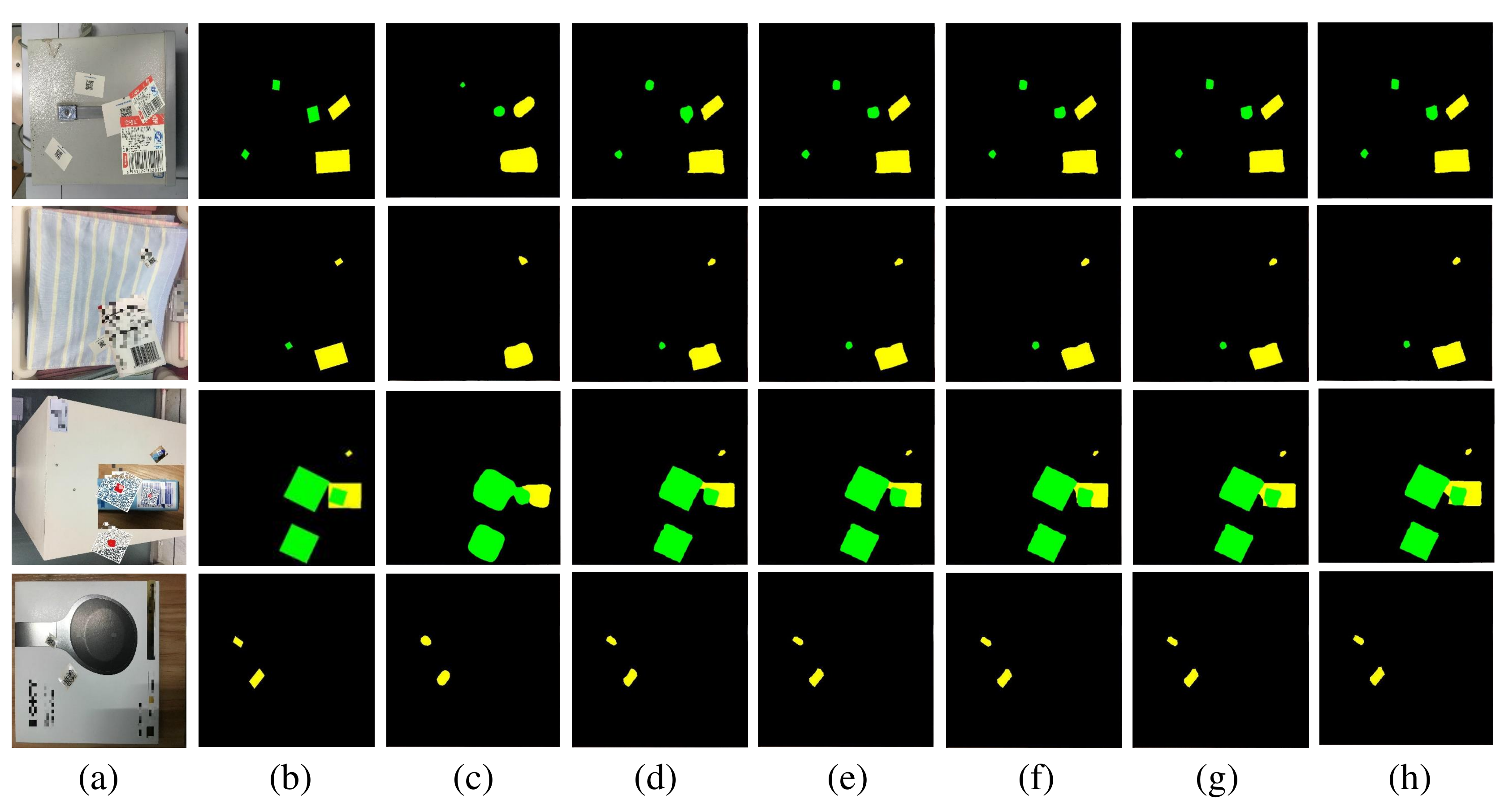}
\caption{ Visual comparison on Barcode-30k data. (a) Image. (b) Ground Truth. (c) FCN-8s [26]. (d) PSPNet[24]. (e) PSPNet-P3M. (f) PSPNet-P3M-PRM(Res5). (g) PSPNet-P3M-PRM(Res1-Avgpooling). (h) BarcodeNet(Res1-2Convs).}
\label{fig:results}
\end{figure}

More visualization comparisons between BarcodeNet and other methods are shown in Fig~\ref{fig:results}. 
Our baseline model FCN-8s has two main problems. One is that it missed a lot of tiny barcodes. Another is the blurred boundaries, both leading to poor segmentation results.

Then PSPNet gets a large improvement, shown in the third column of Fig.5. However, the boundaries and corners are still not good enough. When further replacing P2M with P3M module, connecting PRM from Res5, change to connecting PRM from Res1 by average pooling, and change to connecting PRM from Res1 by two Convs, the results are gradually more accurate, and the problem that PSPNet suffers from has been greatly relieved. 

In addition, segmentation results of real images are shown in Fig 5. We select challenging cases for visualization of our powerful BarcodeNet. Note that we use the BarcodeNet pre-trained on full Barcode-30k training set.

\begin{figure}[t]
\centering
\includegraphics[width=10cm,height=3.5cm]{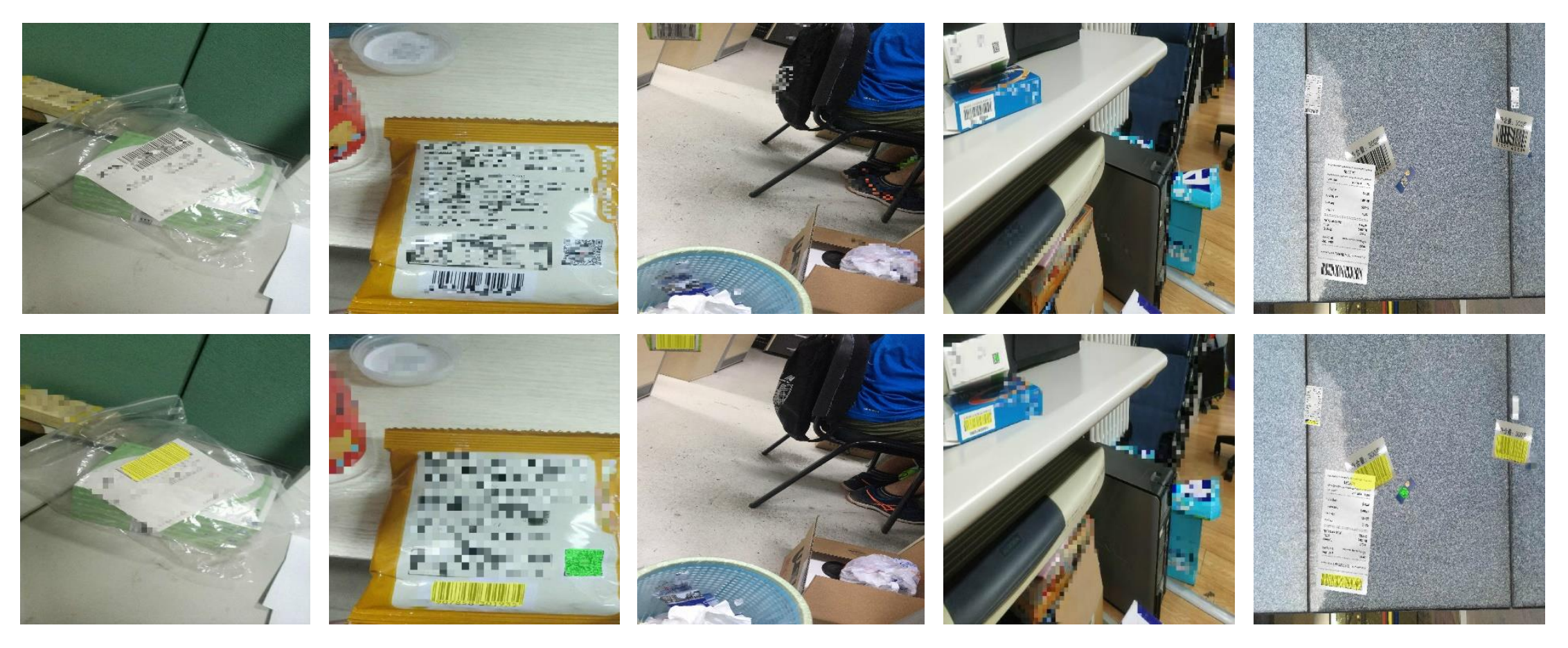}
\caption{Segmentation results of real images collected in office places.}
\label{fig:visual}
\end{figure}


\section{Discussion}
\subsection{Why barcode segmentation, not detection?}
Segmentation task and the detection task for digital signs are very different. For one thing, on the conditional that they are small, dense, overlapped, and randomly cut, it will be hard to detect. Some sampled false positive results are shown in Fig.6. Detection results are computed by OpenCV.

For another, segmentation results support post-processing steps for extracting more useful information, e.g., using the maximum outside rectangle to shot the barcodes, getting target areas with different shapes, not only limited to rectangle targets. However, detection results of specific bounding boxes have more limitations.




  \subsection{Generalization of our method}
Though we mainly focus on barcode images in this paper, note that the synthesis strategy and the network BarcodeNet can be well extended for other targets with similar properties, such as LOGOs, traffic sighs, billboards segmentation, etc. Our synthesis strategy requires little annotations cost, and it provides guidance for time-critical segmentation tasks. 


\begin{figure}
\centering
\includegraphics[width=9cm,height=7cm]{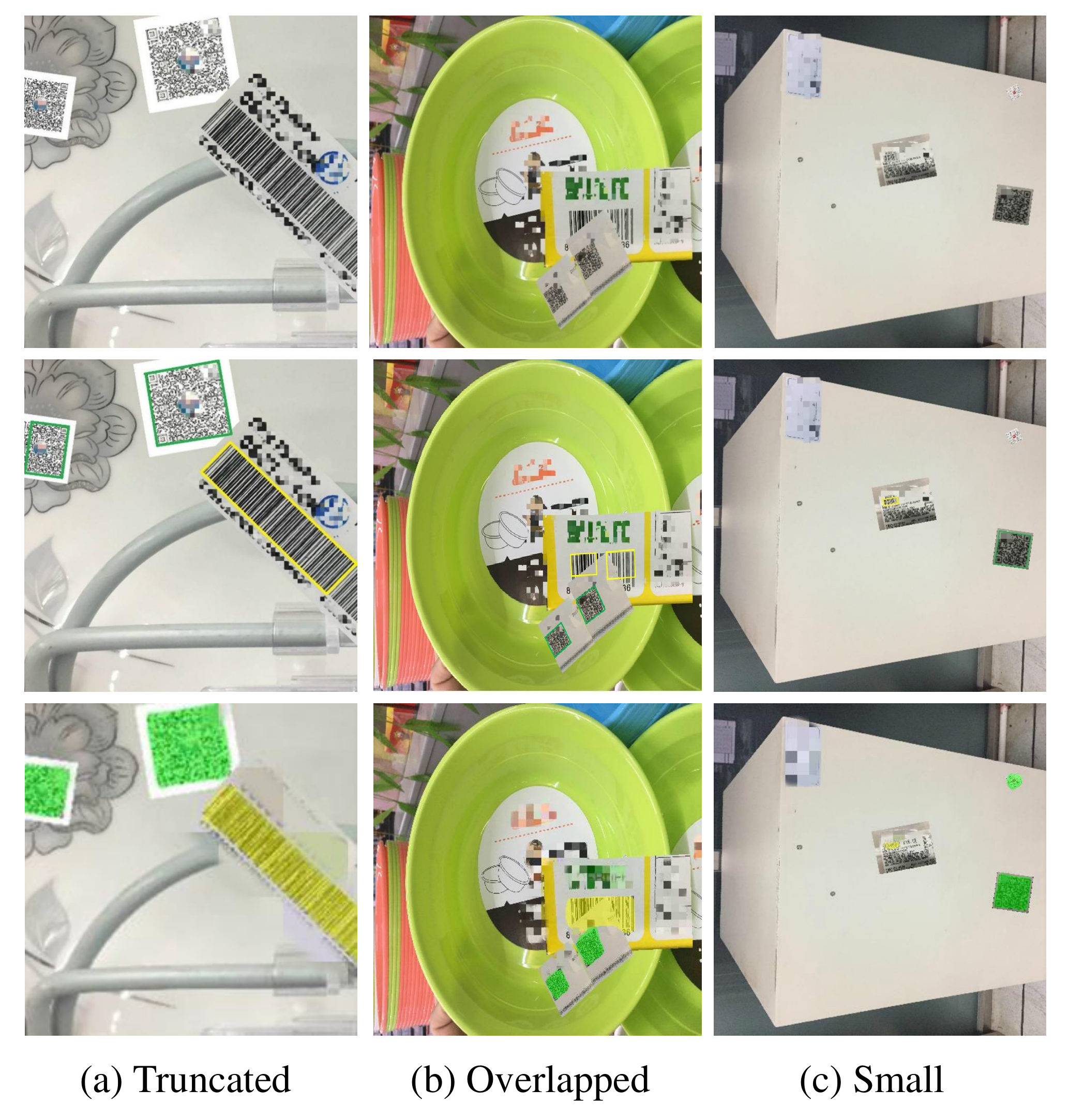}
\caption{Compare qualitative results between barcode detection and segmentations in situations of (a) truncated, (b) overlapped and (c) small objects.}
\label{fig:detection_segmentation}
\end{figure}

\section{Conclusion}
In this paper, we firstly propose a method for barcode image synthesis, by which we introduce a large-scale barcode image dataset, Barcode-30k. Furthermore, according to the appearance characteristics of digital signs, we introduce a deep Dual Pyramid Network - BarcodeNet. It reached a high mIoU of 0.9536 on the Barcode-30k validation set. Also, when applied to segment real images, the BarcodeNet pre-trained on Barcode-30k can get great segmentation results. What's more, this method and model can be extended to the semantic segmentation of other objects with regular shapes.


%
%
%
\bibliographystyle{splncs04}
\bibliography{detection}
\end{document}